# Enhancing Prediction Models for One-Year Mortality in Patients with Acute Myocardial Infarction and Post Myocardial Infarction Syndrome

Seyedeh Neelufar Payrovnaziri[a], Laura A. Barrett[a], Daniel Bis[b], Jiang Bian[c], Zhe He[a]

[a] School of Information, Florida State University, Tallahassee, Florida, USA,
[b] Department of Computer Science, Florida State University, Tallahassee, Florida, USA
[c] Department of Health Outcomes and Biomedical Informatics, University of Florida, Gainesville, Florida, USA

**Abstract**

*Predicting the risk of mortality for patients with acute myocardial infarction (AMI) using electronic health records (EHRs) data can help identify risky patients who might need more tailored care. In our previous work, we built computational models to predict one-year mortality of patients admitted to an intensive care unit (ICU) with AMI or post myocardial infarction syndrome. Our prior work only used the structured clinical data from MIMIC-III, a publicly available ICU clinical database. In this study, we enhanced our work by adding the word embedding features from free-text discharge summaries. Using a richer set of features resulted in significant improvement in the performance of our deep learning models. The average accuracy of our deep learning models was 92.89% and the average F-measure was 0.928. We further reported the impact of different combinations of features extracted from structured and/or unstructured data on the performance of the deep learning models.*

*Keywords:*

Electronic Health Records, Machine Learning, Deep Learning

## Introduction

In 2016, the top two death causes were heart disease and cancer, accounting for 44.9% of all deaths in that year [1]. Based on a recent report from the American Heart Association, cardiovascular disease and stroke are accounted for tremendous economic and health-related burdens in the United States and worldwide [2]. Acute myocardial infarction (AMI) is an event of myocardial necrosis caused by the unstable ischemic syndrome. It is the leading cause of mortality worldwide [3]. Appropriate management of AMI and timely interventions play a key role in reducing mortality from cardiovascular diseases. Nevertheless, this requires us to understand the past trends and patterns of AMI-related mortality and subsequently to inform the design of future tailored interventions based on the available data and models [4][5].

Prediction models have been increasingly used in hospital settings to assist with risk prediction, prognosis, diagnosis, and treatment planning, ultimately leading to better health outcomes for patients. For example, predictive modeling can inform personalized care based upon health conditions of each individual patient [6]. Specifically, mortality prediction models estimate the probability of death for a group of patients based on their characteristics including the severity of their illness and many other associated risk factors for death [7]. They are important complementary tools to assist in clinical decision-making [8][9]. In current clinical practice, score-based mortality prediction systems, such as the series of the acute physiology and chronic health evaluation (APACHE) scoring system, are widely used to help determine the treatment or medicine should be given to patients admitted into intensive care units (ICUs) [10]. Nevertheless, these scoring systems have significant limitations, e.g., 1) they are often restricted to only few predictors; 2) they have poor generalizability and may be less precise when applied to specific subpopulations other than the original population used for the initial development; and 3) they need to be periodically recalibrated to reflect changes in clinical practice and patient demographics [6]. The wide adoption of electronic health record (EHR) systems in healthcare organizations allows the collection of rich clinical data from a huge number of patients [11]. Large EHR data enables one to 1) build more precise prediction models considering a wider range of patient characteristics; 2) be able to refresh these prediction models more frequently with less engineering efforts; and 3) improve the quality of these prediction models with fewer issues such as the common generalization problem [12].

One contemporary approach to build these prediction models is to use Machine Learning (ML) methods. ML is a field of computer science closely related to artificial intelligence that has drawn significant attention in the last few years. ML methods can be used to extract patterns and to predict different outcome variables (e.g., mortality) based on a training dataset. They have been shown to improve the predictive power in many real-world prediction tasks; and especially on biomedical problems, ML methods can lead to a better prognosis with richer predictors compared to traditional statistical approaches [6][13]. Most ML methods require significant feature engineering efforts, which rely on a deep understanding of the data and their underlying relationships with the outcome variable. Traditional artificial neural networks, even though relaxed the requirements of feature engineering, have a limited number of layers, connections and learning capacity because of the constraints of their computational power. In recent years, with the fast growing evolutions in both computer hardware (e.g., graphics processing unit, GPU) and training algorithm developments (e.g., the backpropagation algorithm that fine-tunes the whole network toward optimized representations [14]), deep learning systems now have the ability to use multi-layer architecture to learn patterns based on raw input data in every layer, in which features are not engineered by human but are learned from data automatically.

In recent years, a number of studies have deployed different deep learning architectures to predict mortality using EHR data.

For example, Du et al. used a deep belief network (DBN) to predict critical care patient's 28-days mortality [15].

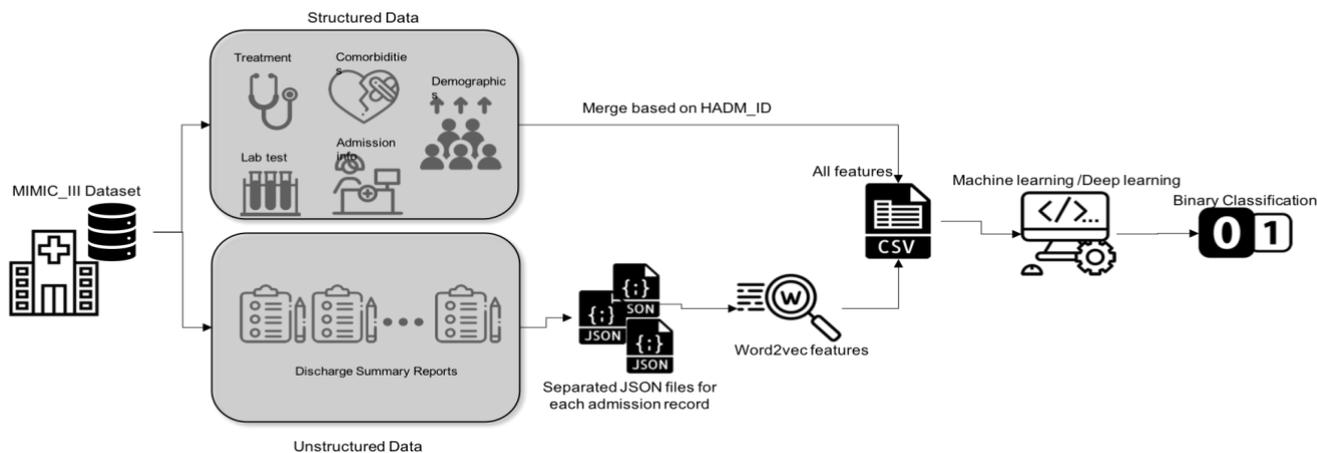

*Figure 1– The workflow of the study (Icons made by https://www.flaticon.com)*

Zahid et al. used self-normalizing neural networks to predict 30-day mortality and hospital mortality in ICU patients [16]. Rajkomar et al. proposed a new representation of raw medical data and used deep learning to predict multiple medical events including in-hospital mortality 24 hours after admission [17]. However, these studies either did not consider free-text data in their feature sets or were only concentrated on short-term mortality prediction such as 24-hour mortality, for which any interventions might be too late.

In a previous study [18], we built a number of machine learning models using structured EHR data including admission information, demographics, diagnoses, treatments, laboratory tests, and chart values. The aim of the study was to predict one-year mortality in patients diagnosed with AMI or PMI. We compared the prediction results of these different machine learning models (i.e., shallow learners such as random forest and adaboost); and then compared the prediction performance of the best performing shallow learners to a deep learning model—a fully connected neural network. The results showed that the deep learning model enhanced recall and F-measure metrics (i.e., from a recall of 0.744 to 0.820; and a F-measure of 0.715 to 0.813) while preserving a good prediction accuracy of 82.02%.

In this study, we advance our previous work by adding unstructured data to the previous models. Word embedding features are extracted from free-text discharge summaries and added to the structured features. This study aims to improve the deep learning model performance using the mixture of both structured and unstructured data, which will be called mixed data throughout this paper. Also, the best performing shallow learners from the previous study are compared once again with the deep learning model using the same mixed data. Further, we examine the performance of the deep learning model using the unstructured set of data only, as well as five different combinations of the structured and unstructured data. We aim to determine which set of features contributes the most in enhancing the performance of deep learning models.

## Methods

In this section, we first briefly introduce our preparation of the structured data as well as the free-text data. Our goal was to build and compare deep learning and traditional machine learning (i.e., shallow learner) models to predict one-year mortality in ICU patients with AMI and PMI. Many tasks in natural language processing (NLP) have benefited from neural word representations. These representations do not treat words as symbols; but rather can capture the semantics of the words and reflect their semantic similarities. These methods that represent words as dense vectors are referred to as "neural embeddings" or "word embeddings". Word embeddings have been proven to benefit a variety of NLP tasks [19]. Then, we briefly introduce the best performing shallow learners from the previous study, which was used to build new models based on the new mixed dataset. Then, we explain the architecture of the deep learning model. A workflow of this study is depicted in Figure 1.

**Dataset Processing**

*Data Source and Patient Cohort*

We used the data from the Medical Information Mart for Intensive Care III (MIMIC-III). MIMIC-III is a freely accessible, de-identified critical care patient database developed by the MIT Lab for Computational Physiology [20]. The latest version of the MIMIC-III dataset includes information about 58,000 admissions to the Beth Israel Deaconess Medical Center in Boston, Massachusetts from 2001 to 2012. Using the International Classification of Diseases, Ninth Revision (ICD-9) codes of 410.0-411.0 (Acute myocardial infarction, Postmyocardial infarction syndrome), we identified 5,436 admissions into our experiment dataset.

*Structured Data Processing*

The structured data in MIMIC-III include admission information (e.g., total days of admission, initial emergency room diagnoses, etc.), demographics (e.g., age at admission, gender, etc.), treatment information (e.g., cardiac catheterization, cardiac defibrillator, and heart assist anomaly, etc.), comorbidity information (e.g., cancer, endocrinology, etc.) and lab and chart values (e.g., cholesterol ratio, alanine transaminase, etc.). We selected these features based on the features used in similar studies. For details, see [18]. They were further refined and limited by their availabilities in MIMIC-III. To ensure that there was only one admission per instance, duplicates were removed. If duplicates existed because of multiple treatments or comorbidities for the same admission, all of them were counted. Regarding the demographics, since age and death age for people over 89 years old were masked in MIMIC-III by adding 211 to the actual age, we changed them back by subtracting 211 from their value. Some lab values were entered with a '0' and associated with a note of 'see comment'.

Thus, 0 values were removed from the lab. Also, the lab or chart values that were biologically invalid were removed. We replaced removed values with the mean value of each feature column. The data was imbalanced with 30% positive and 70% negative cases. The outliers were removed based on the interquartile range rule [18]. Data values were normalized between 0 and 1.

*Unstructured Data Processing*

The unstructured data were retrieved based on the corresponding admission IDs in the structured dataset using NOTEEVENTS table of MIMIC-III, from discharge summaries associated with each admission. Discharge summaries are the main method to communicate a patient's plan of care to the next provider [21]. Thus they include rich information about a patient's condition and treatments. Skip-gram model is a neural embedding method to learn an efficient vector representations of words from unstructured text data. These representations of words encode many linguistic regularities and patterns. The Skip-gram model finds the word representations that can predict the surrounding words [22]. The resulting dense vectors are called word embeddings. In this work, we opted to use document embeddings which is the average of word embeddings vectors for the words in the discharge summary of an admission, because the order of the words is not associated with the outcome (i.e., mortality). We used the Word2Vec algorithm in the Gensim library, a free Python library for processing plain text [23]. We used the embeddings pre-trained with Gensim using scientific articles (i.e., PubMed abstracts and PubMed Central full texts [24]).

*Case Labeling*

The goal of this study was to predict one-year mortality, i.e., whether a patient will die within a year after admission or not. Thus, the admission records of the patients who died within a year were labelled as positive instances, and of those who did not die within a year were labelled as negative instances.

**Predictive Modeling**

*Machine Learning Models*

Waikato Environment for Knowledge Analysis (WEKA) is a freely available Java-based software developed at the University of Waikato, New Zealand. Based on the results from our previous study [18], simple logistic and logistic model trees (LMT) classifiers in WEKA produced the best results using the dataset of structured features including admission information, demographics, treatment information, comorbidity information, lab values, and chart values. The simple logistic classifier in WEKA, builds linear logistic regression models. LMT in WEKA builds classification trees with logistic regression functions at the leaves [25].

*Deep Learning Model*

The deep learning model we used in this work consists of four layers (i.e., the input layer, two hidden layers, the classification layer). Figure 2 shows the deep neural network architecture used in this study. We used the Keras library [26] running on top of the Tensorflow framework [24], as well as a number of other Python packages including SciPy [27], Scikit-learn [28], NumPy [29], and Pandas [30].

The deep neural network we used for this study had 2 hidden layers fully connected with 400 neurons in each layer. The input dimension was 279. We used hyperbolic tangent activation function in hidden layers, and softmax activation function in the classification layer. We used the stochastic gradient descent method for optimization and categorical cross entropy as the loss function. To avoid over-fitting, we used L2 regularization in each hidden layer as well as dropout with a rate of 0.3. Batch size was 100 and epoch size was 60. In each hidden layer we applied batch normalization. All the deep learning architecture settings were chosen based on an extensive examination of different values and their impact on the overal performance. Since the data size was limited, we considered 10-fold cross-validation technique for model validation. We shuffled the data before each run.

**Model Evaluation**

We ran each algorithm 10 times. In each run, the data was shuffled randomly and 10-fold-cross-validation was employed to evaluate the performance (90% for training and 10% for testing). The performance metrics (i.e., accuracy, precision, recall and F-measure) were averaged after 10 folds.

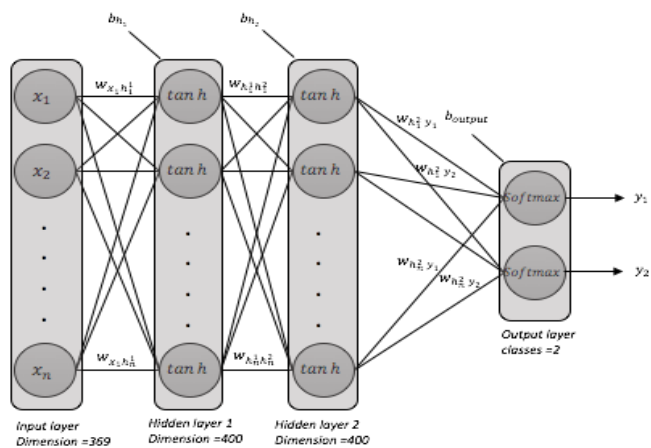

*Figure 2 – a fully connected deep neural network architecture: two hidden layers, each with 400 neurons, initial weights=random uniform, initial bias=zeros, learning rate =0.001*

The accuracy metric reports the model overall performance on the test set; however, recall and precision metrics of these models are more important in our task. If the actual outcome for a patient is mortality within a year, recall metric evaluates how many times the model was able to predict this correctly that a patient died within a year (true positive) out of all the patients who actually died within a year (true positive + false negative). Precision, on the other hand, evaluates how many times a correct prediction (true positive) happened out of all positive predictions made by the model regardless of their correctness (true positive + false positive). False negative in this study means that a patient who is predicted to live within a year actually died. False positive in this study means a patient who is predicted to have died within a year did not die. F-measure evaluates the balance between these two metrics. Although the receiver operator characteristics (ROC) curve is another popular evaluation metric, its interpretation requires caution when used with imbalanced datasets [31]. Since our dataset is imbalanced, we used precision-recall plot for the visual evaluation of the binary classifier.

**Results**

In our previous study [18], we first compared the performance of various machine learning models on each set of structured features separately and then compared them to the performance of machine learning models on the combined dataset (admission + treatment + lab and chart values + demographics + comorbidities). We observed that LMT and simple logistic models achieved the best accuracy of 85.12% on the combined dataset. The recall values were low (from 0.499 to 0.660). Only the J48 classifier yielded a precision of 0.993 using the admission dataset alone, while other performance metrics decreased notably comparing to using the combined dataset. Then we showed that a deep learning model can enhance the performance. Our deep learning model achieved 82.02% accuracy, while boosted recall and F-measure metrics to 0.820 and 0.813, respectively. All features used in the previous study were derived from structured data.

*Table 1– Comparing machine learning models to deep learning model based on the mixed dataset*

| Model | Accuracy | Precision | Recall | F-measure |
|---|---|---|---|---|
| LMT | 85.78% | 0.856 | 0.621 | 0.724 |
| Simple Logistic | 85.71% | 0.863 | 0.623 | 0.723 |
| Deep Learning | **92.89%** | **0.931** | **0.929** | **0.928** |

In this work, we first compared the performance of machine learning and deep learning models on the mixed dataset (i.e., features from both structured and unstructured data). Then, we created different combinations of structured data with unstructured data to examine which set of features has more predictive power for our classification task. Table 1 shows the performance of the two top performing traditional machine learning models (as obtained from our previous study) and a deep learning model on the mixed dataset. The deep learning model outperformed the best shallow learners considerably.

In Table 1, we can see that the precision values of shallow learners are higher than their recall values, which means they are exact but not complete. A low recall value indicates a large number of false negatives (i.e., incorrectly classified as not dying within a year), which is suboptimal in this classification task. The dimension of data in our previous study was 79 considering only features from structured data. Adding features derived from unstructured data increased the total number of features and increased the input data dimension up to 279. Table 2 illustrates the comparison between the previous work and current study. We can see from the results that shallow learners did not benefit from more features (and higher data dimensionality). Accuracy slightly improved, while precision slightly dropped. Recall improved less than 0.03. Unlike the shallow learners, our deep learning model showed considerable improvements with more than 10% increase in accuracy and ~10% improvement in both precision and recall. Further, we were interested in comparing the performance of deep learning models using only free-text features vs. using different combinations of structured and free-text features. Results are summarized in Table 3.

From the results we observed, demographic and admission information are two key groups of structured features in enhancing the deep learning model. Demographic information in this dataset includes age at admission, gender, religion, ethnicity and marital status. Admission information includes total days of admission, discharge location and initial ER diagnosis as AMI or rule out AMI.

*Table 2– Comparing machine learning and deep learning models based on structured dataset vs. mixed dataset (structured + unstructured),-p means previous study, -c means current study*

| Model | Accuracy | Precision | Recall | F-measure |
|---|---|---|---|---|
| LMT-p | 85.12% | 0.867 | 0.594 | 0.705 |
| LMT-c | 85.78% | 0.865 | 0.621 | 0.724 |
| Simple Logistic-p | 85.12% | 0.867 | 0.549 | 0.705 |
| Simple Logistic-c | 85.71% | 0.863 | 0.623 | 0.723 |
| Deep Learning-p | 82.02% | 0.831 | 0.820 | 0.813 |
| Deep Leanirng-c | 92.89% | 0.931 | 0.929 | 0.928 |

*Table 3– Comparing using unstructured data only vs. different combinations with structured data in the deep learning model*

| Data | Accuracy | Precision | Recall | F-measure |
|---|---|---|---|---|
| Free text | 81.83% | 0.836 | 0.818 | 0.816 |
| Free text + lab results | 83.61% | 0.853 | 0.836 | 0.833 |
| Free text + treatment | 84.15% | 0.850 | 0.841 | 0.840 |
| Free text + comorbidity | 84.69% | 0.856 | 0.846 | 0.842 |
| Free text + demographics | 87.37% | 0.881 | 0.874 | 0.872 |
| Free text + admissions | 88.25% | 0.885 | 0.882 | 0.881 |

The combination of admission information with free-text features produced an accuracy of 88.25% in the deep learning model; while, the accuracy of the deep learning model based on the combination of demographics data with free-text features was 87.37%. We compared these two models to the accuracy of another deep learning model based on the complete mixed dataset, which produced an accuracy of 92.89%. Figure 3 illustrates the precision-recall curve resulted after 10 rounds of deep learning algorithm run. Table 4 shows a comparison of other recent works in mortality prediction using deep learning methods on EHR data.

*Table 4– Comparing recent work in mortality prediction using deep learning methods on EHR data*

| Paper | Mortality Prediction Task | AUC | ACC |
|---|---|---|---|
| Payrovnaziri et al. *(this paper)* | 1-year | 0.916 | 92.89% |
| Du et al.[15] | 28-days | Not reported | 86% |
| Zahid et al.[16] | 30-days/hospital | 0.8445/0.86 | Not reported |
| Rajkomar et al.[17] | 24 h after admission | 0.92-0.94 | Not reported |

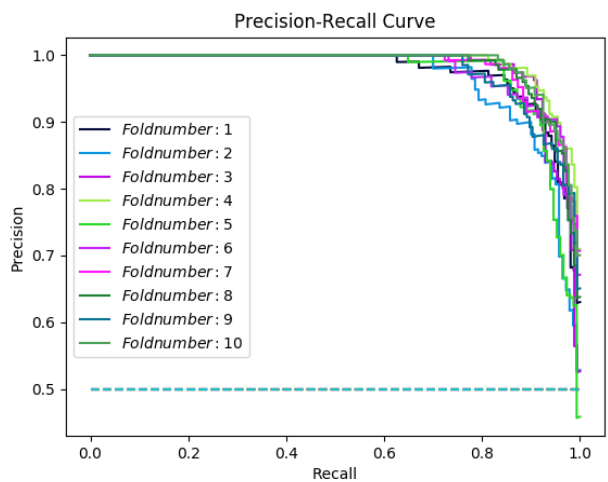

*Figure 3– Precision-Recall Curve, after 10 runs average precision = 0.931, average recall = 0.929*

## Discussion and Conclusions

In this work, we enhanced our previous deep learning model by combining unstructured and structured data to predict one-year mortality in ICU patients with AMI and PMI. For unstructured data, we extracted word embedding features from discharge summaries of each patient admission. While these word embedding features had no impact on the shallow learners, the performance of our deep learning model increased and achieved an accuracy of 92.89%, precision of 0.931, recall of 0.929 and F-measure of 0.928.

Our findings suggest that a richer data dimension through adding features from unstructured data will enhance deep learning model performance. We also confirmed our previous findings that initial emergency room diagnosis, gender, age, and ethnicity are important factors for the prediction of one-year mortality. One limitation worth noting is that using ICD-9 CM codes for cohort identification may introduce some noise. But this noise should not impact the findings of this study. In future work, we are interested in: 1) designing deep neural network ensembles that have the potential to further improve the model performance; 2) exploring the unstructured sequential data through other state-of-the-art models such as recurrent neural networks and long short-term memory (LSTM) techniques; and 3) the potential to enrich the textual features by extracting Unified Medical Language System (UMLS) concepts from the free-text data.

**Address for correspondence**

Zhe He, PhD. School of Information, Florida State University, 142 Collegiate Loop, Tallahassee, Florida, USA. Email: zhe@fsu.edu